\documentclass{article}

\usepackage{enumitem}
\usepackage{verbatim}
\usepackage{PRIMEarxiv}
\usepackage{enumitem}
\usepackage{verbatim}
\usepackage{fancyhdr}
\usepackage[utf8]{inputenc}
\usepackage[T1]{fontenc}   
\usepackage{hyperref}       
\usepackage{url}            
\usepackage{booktabs}      
\usepackage{amsfonts} 
\usepackage{nicefrac}       
\usepackage{microtype}      
\usepackage{lipsum}
\usepackage{fancyhdr}       
\usepackage{graphicx}      
\usepackage{multirow}
\usepackage{graphicx}
\usepackage{textcomp}
\usepackage[export]{adjustbox}
\graphicspath{{media/}}    

\title{Soft Gripping: Specifying for Trustworthiness}

\author{
   Dhaminda B. Abeywickrama\\
   Department of Computer Science \\
   University of Bristol, UK \\
   \texttt{dhaminda.abeywickrama@bristol.ac.uk}
   \And 
	Nguyen Hao Le \\
	Department of Engineering Mathematics \\
	University of Bristol, UK \\
	\texttt{anthony.le@bristol.ac.uk} 
   \And
    Greg Chance \\
	Department of Computer Science \\
	University of Bristol, UK \\
	\texttt{greg.chance@bristol.ac.uk} 
   \And
   Peter D. Winter \\
   School of Sociology \\
	University of Bristol, UK \\
   \texttt{peter.winter@bristol.ac.uk} 
   \And
   Arianna Manzini \\
   Bristol Medical School\\
	University of Bristol, UK \\
   \texttt{amanzini.ethics@gmail.com} 
   \And 
	Alix J. Partridge \\
	Department of Mechanical Engineering \\
	University of Bristol, UK \\
	\texttt{alix.partridge@bristol.ac.uk} 
	\And
	Jonathan Ives \\
	Bristol Medical School\\
	University of Bristol, UK \\
	\texttt{j.ives@bristol.ac.uk} 
   \And
	John Downer \\
	School of Sociology \\
	University of Bristol, UK \\
	\texttt{john.downer@bristol.ac.uk} 
	\And 
   Graham Deacon \\
   Robotics Research Team\\
    Ocado Technology, UK \\
   \texttt{graham.deacon@ocado.com} 
   \And
	Jonathan Rossiter \\
	Department of Engineering Mathematics \\
	University of Bristol, UK \\
	\texttt{jonathan.rossiter@bristol.ac.uk}   
   \And
   Kerstin Eder \\
   Department of Computer Science \\
   University of Bristol, UK \\
   \texttt{kerstin.eder@bristol.ac.uk} 
	\And
Shane Windsor \\
Department of Aerospace Engineering \\
University of Bristol, UK \\
\texttt{shane.windsor@bristol.ac.uk} 
}

\begin{document}
\maketitle
\begin{abstract}
	Soft robotics is an emerging technology in which engineers create flexible devices for use in a variety of applications. In order to advance the wide adoption of soft robots, ensuring their trustworthiness is essential; if soft robots are not trusted, they will not be used to their full potential. In order to demonstrate trustworthiness, a \emph{specification} needs to be formulated to define what is trustworthy. However, even for soft robotic grippers, which is one of the most mature areas in soft robotics, the soft robotics community has so far given very little attention to formulating specifications. In this work, we discuss the importance of developing specifications during development of soft robotic systems, and present an extensive example specification for a soft gripper for pick-and-place tasks for grocery items. The proposed specification covers both functional and non-functional requirements, such as reliability, safety, adaptability, predictability, ethics, and regulations. We also highlight the need to promote \emph{verifiability} as a first-class objective in the design of a soft gripper.	
\end{abstract}
\newpage
	\section{Introduction}\label{introduction}

Soft robotics is an emerging technology in which engineers create flexible devices for use in a variety of applications, such as surgery, prosthetics, space exploration, and grocery picking. 
In order to advance the wide adoption of soft robots, ensuring their trustworthiness is essential. 
\emph{Trust} can be earned and lost over time. Trust can be defined by various research disciplines in different ways; in comparison something is \emph{trustworthy} when it is deserving of trust. 

Autonomous systems (AS) (e.g. soft robotic systems) can be \emph{trustworthy} when their design, engineering, and operation generate positive outcomes and mitigate harmful outcomes~\cite{Naiseh2022}.  
The trustworthiness of AS depends on many factors like (i) robustness in uncertain and dynamic environments; (ii) accountability, explainability, and understandability to different users; (iii) assurance of their design and operation through verification and validation (V\&V) activities; (iv) confidence to adapt their functionality; (v) security to counter attacks; (vi) ethics and human values in their use and deployment; and (vii) governance and regulation of their design and operation~\cite{Naiseh2022,Abeywickrama2022}.  
Different techniques can be used to demonstrate the trustworthiness of a system, such as formal verification at design-time, monitoring and runtime verification, synthesis, and test-based methods \cite{Abeywickrama2022}. 
However, common to all these techniques is the need to formulate \emph{specifications}. 
A \emph{specification} is a detailed formulation that provides ``a definitive description of a system for the purpose of developing or validating the system''~\cite{ISO24765:2017}. 

\emph{Soft robotic grippers} for manipulating objects are considered to be one of the most mature areas in soft robotics. 
However, even for this relatively well advanced application area the soft robotics community has so far given very little attention to formulating specifications (e.g. \cite{Shi2023,Cheng2021,Liu2021,Chen2018,Cai2021,Hwang2020,Shin2021}). Creating a specification as part of a development process is an important step in which the requirements for the system are agreed and defined precisely. 
The specification then provides a benchmark against which to assess different design options. A well-defined specification also provides a set of criteria against which to verify the performance of the system during design and in operation.
 
In addition to formulating a well-defined specification, a key technique which can explored to improve the trustworthiness of a soft robotic gripper is \emph{verifiability}. 
Verifiability considers verification as an integral part of the system specification and the system design \cite{Mousavi2022}, where it can be promoted to a primary system design objective \cite{Eder2021}. 
Verifiability will essentially give rise to systems, which by their construction, deserve our trust~\cite{Mousavi2022}. 
The main contributions of this paper are as follows:
\begin{itemize}
	\item We provide an extensive example specification to ensure the trustworthiness of a soft robotic gripper \cite{Partridge2022} with both functional and non-functional requirements, such as reliability, safety, adaptability, predictability, ethics, and regulations.
	\item We highlight the importance of promoting \emph{verifiability} as a first-class objective in the design of a soft robotic gripper. Also, we provide illustrations of how to formulate verifiable requirements for soft grippers.
\end{itemize}
We explore this novel topic using a case study of pick-and-place tasks of grocery items \cite{Triantafyllou2019, Sotiropoulos2018} involving a recycled soft gripper \cite{Partridge2022}. 

The rest of the paper is organised as follows.  
In Section~\ref{background-relatedwork}, we provide background information to this work, key related works, and a brief description of the case study.  
Section~\ref{specification-gripper} provides a detailed specification of a soft gripper, and in Section~\ref{verifiability}, we highlight the significance of promoting verifiability.
Finally, Section~\ref{summary-conclusions} concludes the paper. 	

\section{Background, Related Work and Case Study}\label{background-relatedwork}

\subsection{Background}\label{background}
\subsubsection{Recycled Soft Gripper}
\begin{figure}
	\centering
	\includegraphics[width=0.5\textwidth]{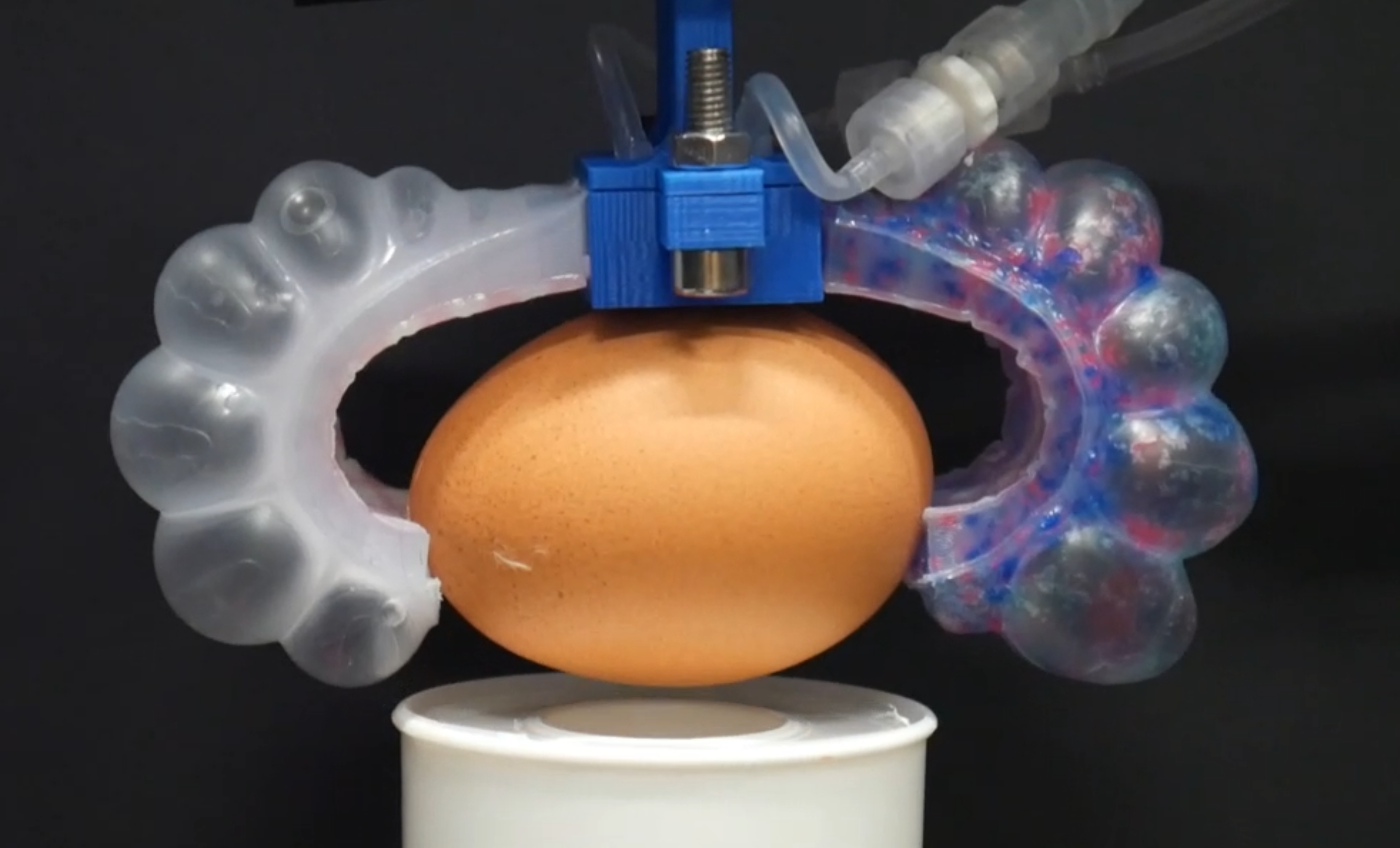}
	\caption{A soft pneumatic gripper with a pristine silicone finger and a second finger containing 30\% recycle silicone material grasping a fresh egg~\cite{Partridge2022}.}
	\label{gripper}
\end{figure}

The soft robotic gripper used in this work is a two-finger fluidic elastomer actuator, measuring 12 x 134 x 6 mm \cite{Partridge2022} (see Fig.\ref{gripper}). The gripper is fabricated using a two-part moulding process with a fabric-constraining layer and comprises a mix of 70\% pristine EcoFlex 00-30 silicone elastomer and 30\% recycled EcoFlex 00-30 granules that are 1 mm to 2 mm in size.

The operating procedure for the gripper is simple. To actuate the gripper, a positive pressure is introduced to the system that acts to inflate chambers within the gripper body. As the chambers expand, the constraining layer on the base of the gripper prevents expansion of the base of each chamber, while the top of each chamber is free to expand. This differential expansion of top and bottom results in a curving of each chamber, which then leads to an overall curling of the gripper fingers to grasp an object. When vented, this pressure is removed, and the gripper fingers return to a flattened state. 

\subsubsection{Standards} 
Although no direct industry standards have been defined for soft grippers so far, there are several standards in the area of rigid robotics that provide: (i) a set of terminology and definitions for robotic grippers (ISO 14539:2000) and hands \cite{Falco2018}); and (ii) guidance related to the safe spaces, speeds, and forces that a gripping system needs to function (ISO 10218-1/10218-2, and ISO/TS 15066). In 2022, ASTM International (American Society for Testing and Materials) formed a new subcommittee (F45.05) with the aim of developing new standards for evaluating performance in various areas of soft robotic manipulation.

\subsection{Related Work}\label{relatedwork}
Most existing approaches (e.g. \cite{Netzev2023,Hong2022,Bhattacharya2019,Tadakuma2020,Loh2014,Nishikawa2019,Mohan2020}) only describe some technical requirements and parameters of the end-effector to drive its physical design, such as its dimensions, weight, and material properties. 
For example, Netzev et al.~\cite{Netzev2023} list several technical requirements that are essential for achieving the desired result of their grasping method, such as dimensions and heights of the largest and smallest objects which can be gripped, maximum object weight, and gripping force. 
Also, most specifications provided for soft grippers only describe their functional requirements (e.g. \cite{Shi2023}), and only a few approaches describe non-functional requirements, such as \emph{adaptability} and \emph{performance} (e.g. \cite{Cheng2021,Liu2021,Chen2018,Cai2021,Hwang2020,Shin2021}). 
For example, Shi el al. \cite{Shi2023} summarise functional requirements to characterise soft robots with respect to their force, dynamics, and stiffness identification. 
Cheng et al~\cite{Cheng2021} conduct a series of static and dynamic gripping tests that use different forces and fingertip displacements, and demonstrate enhancements to \emph{adaptability} and \emph{performance} of their three-finger soft-rigid gripper. 
However, these works do not cover a wide range of properties which affect the trustworthiness of a soft gripper, as proposed in this work. 

\subsection{Case Study: Pick-and-Place Tasks of Grocery Items}
Let us consider an automated warehouse, where items are picked from storage crates using a soft robotic gripper and are then placed in delivery crates for delivery. 
An example grocery use case is that of Ocado, which is considered the world's largest online-only supermarket \cite{Triantafyllou2019, Sotiropoulos2018}. 
A range of uncertainties make automation of this process challenging: (i) items can vary in their shape, size, packaging, and orientation; (ii) some items are fragile or deformable; (iii) geometrically constrained and relatively cluttered operating environment~\cite{Triantafyllou2019}; (iv) manufacturing inconsistencies (low tolerances) in the gripper; and (v) elastic nature of the materials used in the gripper, which can lead to performance degradation.

In this case study, we consider four \emph{classes} of items: (i) \emph{soft-fragile} items (e.g. cake, bread, strawberry, bayberry), (ii) \emph{soft-non-fragile} items (e.g. dish sponge), (iii) \emph{hard-fragile} items (e.g. light bulb, egg), (iv) \emph{hard-non-fragile} items (e.g. plastic spoon). 
The objects being picked can be regular-shaped items (e.g. sphere, cube, cone, pyramid, cylinder) or irregular-shaped ones (e.g. strawberries). The robotic pick-and-place task can be structured into a pipeline of four main tasks: (i) pre-grasping, (ii) ascension (grasping), (iii) translation (transport), and (iv) descension (placement). 

\section{Specification of a Soft Gripper}\label{specification-gripper}
\begin{figure*}
	\centering
	\includegraphics[width=1.0\textwidth]{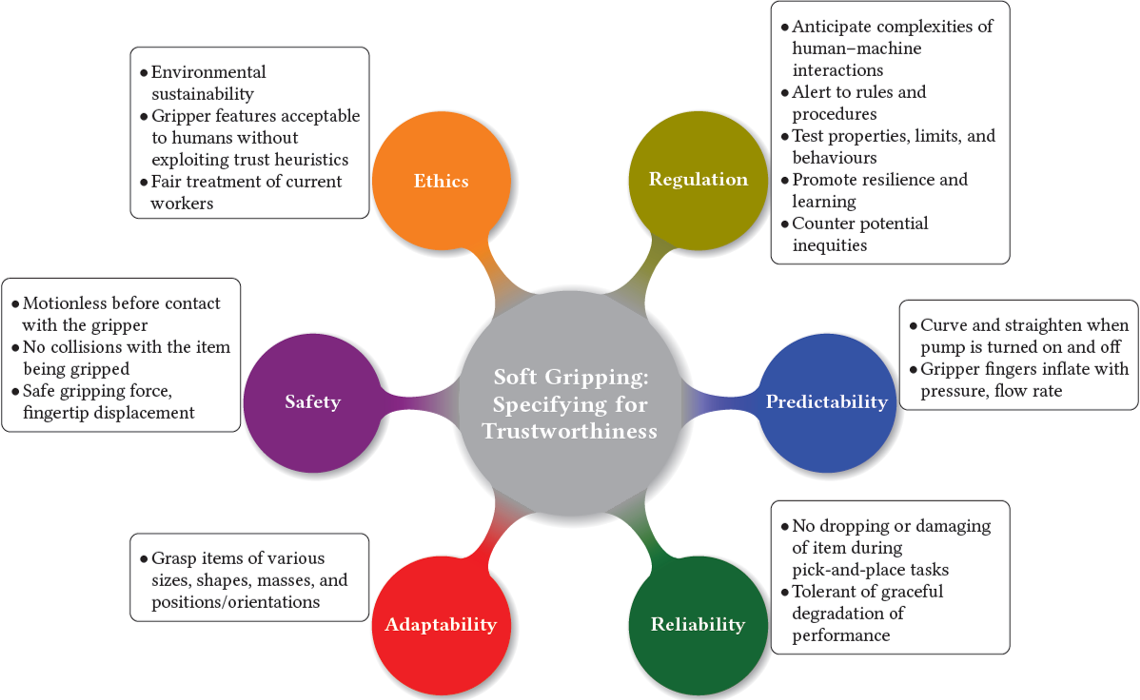}
	\caption{Soft gripper: Specifying for trustworthiness. Core properties (coloured) are evaluated through a range of practical evaluations (boxes). }
	\label{SR-spec}
\end{figure*}

In this section, we present an example of a wide-ranging specification with functional and non-functional properties which affect the trustworthiness of a soft gripper. We define these in terms of predictability, reliability, adaptability, safety, ethics, and regulations (see Fig.\ref{SR-spec}). 

This example specification was developed through consultation with requirements engineers, soft robotic developers, industrial users, ethicists, and sociologists. It is an example of the aspects that may need to be specified for a soft gripper rather than an exhaustive specification for every aspect of the system. 
The specification includes functional requirements -- those that specify behaviour the system shall perform~\cite{ISO24765:2017}; and non-functional requirements -- those that specify not what the system will do but how it will do it (quality attributes)~\cite{ISO24765:2017}. 
The engineering requirements (iii a--d) are formulated as `shall' statements following the guidance on writing good requirements \cite{NASA2007}, and the ethics and the regulatory requirements (iii e--f) are discussed using key frameworks from ethics~\cite{Porter2023} and social science~\cite{Macrae2022}. 

Below we define the set of requirements \{RQx\} for a soft gripper across these six properties.
We define the boundaries to measure the success of gripping an item using the conventional bounds of 95\% for success and 5\% for failure. 
Also, in the following, whenever a performance threshold is given, where possible we aim to provide a reference from a published work/experiment. 

\paragraph{\textbf{Predictability}}\label{predictability}
Predictability is ``a property of interaction concerning the degree of confidence with which a user can determine the effect subsequent task execution will have on the achievement of the goal"~\cite{Abowd1991}. A soft gripper needs to be predictable in its behaviour, so it can build a degree of confidence and trust for the end-user.  
\begin{itemize}
	\item RQ1.1: The fingers of the gripper \emph{shall} curve when inflated (pump turned on).  
	\item RQ1.2: The fingers of the gripper \emph{shall} straighten when deflated (pump turned off).  
	\item RQ1.3: The curvature of a finger \emph{shall} be proportional to its internal pressure. 
	\item RQ1.4: The gripper \emph{shall} grasp, transport, and place an item successfully with a repeatability of $\ge$95\%. 
	\item RQ1.5: The fingers \emph{shall} be inflated with a pressure between 3 and 4 psi pressure range \cite{Partridge2022}. 
	\item RQ1.6: The fingers \emph{shall} be inflated with a flow rate between 2 and 3.2 L/min range \cite{DEWIN2022}.
\end{itemize}

\paragraph{\textbf{Reliability}}\label{reliability}
\emph{Reliability} is described as the ``ability of a system or component to perform its required functions under stated conditions for a specified period of time" \cite{ISO24765:2017}. A soft gripper needs to be reliable by not dropping or damaging items during pick-and-place tasks, and should be tolerant of graceful degradation of performance.

\begin{itemize}
	\item RQ2.1: The gripper \emph{shall} hold the item being gripped without damaging it. 
	\item RQ2.2: The gripper \emph{shall} hold the item being gripped without dropping it for at least 10 seconds 95\% of the time (\cite{Sotiropoulos2018}, p. 652).
	\item RQ2.3: The gripper \emph{shall} successfully maintain grasp during the translation of the gripped item for a maximum velocity and acceleration of 0.03 m/s and 0.15 m/s$^2$ (\cite{Triantafyllou2019}; \cite{Cheng2021}).
	\item RQ2.4: The gripper \emph{shall} successfully grasp when the rate of inflation is in the range of 2-3.2 L/min \cite{DEWIN2022}.
	\item RQ2.5: The gripper \emph{shall} experience $\le$ 5\% increase in the dropping of an item across 100 hours of operation (graceful degradation).
	\item RQ2.6: The gripper \emph{shall} experience $\le$ 5\% increase in the damaging of an item across 100 hours of operation (graceful degradation).
\end{itemize}
\paragraph{\textbf{Adaptability}} \label{adaptability}
\emph{Adaptability} is the ``degree to which a product or system can effectively and efficiently be adapted for different or evolving hardware, software or other operational or usage environments" \cite{ISO24765:2017}. In a soft gripper, \emph{adaptability} is key to grasping objects of different shapes, sizes, masses, and positions or orientations. 
\begin{itemize}
	\item RQ3.1: The gripper \emph{shall} hold items of different sizes up to a maximum of 95\% of the opening width of the two fingers without dropping them for at least 10 seconds 95\% of the time.
	\item RQ3.2: The gripper \emph{shall} hold items of different shapes (e.g. sphere, cube, cone, pyramid, cylinder) without dropping them for at least 10 seconds 95\% of the time.
	\item RQ3.3: The gripper \emph{shall} hold items, which can be of regular or irregular shape (e.g. soft-fragile items like strawberry), without dropping them for at least 10 seconds 95\% of the time.
	\item RQ3.4: The gripper \emph{shall} hold an item independent of its orientation without dropping it for at least 10 seconds 95\% of the time. 	  
\end{itemize}

\paragraph{\textbf{Safety}}\label{safety}
This is described as an ``expectation that a system does not, under defined conditions, lead to a state in which human life, health, property, or the environment is endangered" \cite{ISO24765:2017}. 
In this work, we consider safety from the perspective of the item being gripped where physical damage should be avoided during pick-and-place tasks. 
\begin{itemize}
	\item RQ4.1: The item being gripped \emph{shall} be motionless (to minimise harm) before contact with the gripper. 
	\item RQ4.2: The gripping system \emph{shall} not collide with the item being gripped. 
	\item RQ4.3: The gripping system \emph{shall} only make contact with the item using the gripper.
	\item RQ4.4: When grasping a hard-fragile item (e.g. light bulb, egg), the soft actuator \emph{shall} be inflated until the gripping force does not exceed 2 N (\cite{Cheng2021}).
	\item RQ4.5: When grasping a soft-fragile item like cake or bread, the soft actuator \emph{shall} be inflated until the fingertip displacement does not exceed 3 mm (\cite{Cheng2021}).
	\item RQ4.6: When grasping a soft-fragile item like strawberry or raspberry, the soft actuator \emph{shall} be inflated until the gripping force does not exceed 1 N and the fingertip displacement does not exceed 1 mm (\cite{Cheng2021}, p. 14).
\end{itemize}

\paragraph{\textbf{Ethics}}\label{ethics}
Human values inform the design and use of technologies, and the resulting distribution of benefits and burdens, and those values underpin system trustworthiness. Rather than \emph{describing} whether people \emph{trust} a technology, ethicists are interested in \emph{prescribing} whether they \emph{should trust} it. Below are some key ethical requirements for specifying for trustworthiness in a soft gripper in pick-and-place tasks.
\begin{itemize}
	\item RQ5.1: The gripper \emph{shall} 
	be environmentally sustainable. 
	\item RQ5.2: The gripper \emph{shall} avoid distressing human workers.
	\item RQ5.3: The gripper \emph{shall} avoid exploiting humans' trust heuristics.
	\item RQ5.4: The gripper \emph{shall} accommodate fair treatment of current human workers.
\end{itemize}

The use of a recycled gripper may make its manufacturers or the company deploying it \emph{look like} they care for the environment and so appear more trustworthy, thus increasing public trust in them. However, such a system may not perform as well as, or may degrade more rapidly than, a system made of virgin material, thus making it more likely for items to be dropped or damaged. When the item being gripped is food, the risk of food waste increases, with detrimental implications for the environment. Unless the above reliability requirements are met by design, the use of a recycled gripper could be a form of `greenwashing'~\cite{delmas2011drivers}.    

Moreover, while technical requirements aim to prevent \emph{physical} harm, specifying for trustworthiness should include considerations of potential \emph{psychological} harm \cite{Porter2023}. By replicating a human feature (fingers/hand) and behaviour (gripping) without being human-like enough to convince, a soft gripper could produce an uncanny valley effect \cite{moore2012bayesian} that distresses humans. User studies should identify gripper shapes that are acceptable to humans, whilst aiming for optimal gripping.  

Simultaneously, the biomimetic features of soft grippers risk exploiting humans' trust heuristics, which make them trust things they are already familiar with \cite{Manzini}. As the use of soft robotics technologies can result in many unsafe conditions (e.g. material failure, crushing~\cite{abidi2017intrinsic}), trust in the safety of a soft gripper could be misplaced if it was only based on a sense of familiarity with it. To reduce risks of harm, the robot area should be clearly separated from the space accessible to humans.

\paragraph{\textbf{Regulations}}\label{regulation}
To design successful AS we must formulate specifications from a \emph{sociotechnical} perspective, that is not only regarding the technical features of the system as discussed in III a--d, but also social features of its development and use as fundamentally interrelated. So, in designing a soft gripper for pick-and-place tasks for grocery items, we need to be sensitive to sociotechnical requirements, where the analysis of human, social, cultural, and organisational dynamics can help us think about why and how failures happen with regards to AS. In this context, Macrae's \cite{Macrae2022} `SOTEC' (structural, organisational, technological, epistemic, and cultural) framework is a useful approach for identifying domains of sociotechnical risk in AS. The schema can help inform emergent regulation by identifying often-neglected risks. Below we outline each category and consider its application to the pick-and-place tasks of grocery items.

\emph{Structural} sources of risk arise from interactions between human and nonhuman elements in a system.
In a pick-and-place task, such risks may arise from unanticipated disruptions, such as a human moving a grocery item, confusing the perception system. Regulatory requirements should ensure that the system anticipates and accommodates the complexities of human-machine interactions.

\emph{Organisational} sources of risk emerge when organisational structures, such as rules and expectations, are insensitive to the vagaries of real human behaviour. For example, protocols for checking capabilities required for grasping might presuppose an unrealistic degree of diligence. Regulatory requirements should be alert to the practical dimensions of potential rules and procedures.

\emph{Technological} sources of risk arise from the shortcomings of the system itself. These are the myriad risks that engineers would conventionally consider as part of their work. In the context of the pick-and-place pipeline, there may be concern about recycled materials degrading and shedding particles into food. Regulators should ensure that designers effectively test the properties of the recycled materials, their limits and behaviours.

\emph{Epistemic} sources of risk arise from the inherent indeterminacies of knowledge, which create pockets of ignorance that hide unexpected hazards. 
Such pockets are difficult to mitigate (since we don't know what we don't know) but it can be anticipated that an attempt by the system to grasp an object could fail for unexpected reasons, and tailor regulatory provisions in ways that promote resilience and learning.

\emph{Cultural} sources of risk occur from collective values (beliefs and norms) that frame and influence AS design and operation~\cite{Macrae2022}. 
For instance, if a gripper's grasp is optimised for western food items (e.g. tins), this may mean that it underperforms for eastern food items (e.g. bags), negatively affecting a minority of stakeholders. Regulatory requirements must reflect critically on their underlying values, and work to counter potential inequities. 

\section{Verifiability of a Soft Gripper} \label{verifiability}
In addition to formulating a well-defined specification as discussed in the preceding section, \emph{verifiability} can also be explored to improve the trustworthiness of a soft robotic gripper. 
Verification is the process that can be used to increase confidence in the system's correctness against its specification~\cite{Bergeron2000}. 
One can achieve verifiability by giving consideration to verification early in the process, such as during specification and system design \cite{Mousavi2022}, where it can be promoted to a primary system design objective \cite{Eder2021}. 
A unified and holistic approach to verifiability will give rise to systems, which by their construction, deserve our trust~\cite{Mousavi2022}. 

In order to make a system {\em verifiable\/}, a person or a tool must be able to check its correctness~\cite{ISO24765:2017} in relation to its specification \cite{Abeywickrama2022}. 
The main challenge is in specifying and designing the system in such a way that this process is made as easy and intuitive as possible. 
The specific challenges for AS include~\cite{Abeywickrama2022}: 
(i) capturing and formalising requirements including functionality, performance, safety, security and, beyond these, any additional non-functional requirements purely needed to demonstrate trustworthiness; 
(ii) handling flexibility, adaptation, and learning; and 
(iii) managing the inherent complexity and heterogeneity of both the AS and the environment it operates in. 

Gerson~\cite{Gerson1993} identifies several techniques for ensuring verifiability: bounding the verification task, prioritising effort, ensuring traceability, and breaking down high-level requirements into verifiable portions. 
According to Gerson~\cite{Gerson1993}, formulating requirements for verifiability should go far beyond avoiding negative requirements and including numerical tolerances, but also aim to \emph{design for verifiability}. 
This is because the ultimate purpose of specifications is verifying that the end product exhibits the intended properties. 
This requires the intended properties to be \emph{demonstrable}; that is, knowledge of the end result needs to be attainable. 
In this work, we have considered this early when formulating requirements, by planning and confirming how these properties can be verified subsequently (see Table~\ref{Table:Verifiability}). 
For example, requirements RQ1.1, 1.2, 1.4, 2.1, 2.2, and 2.4 can be verified by \emph{observing} the gripping system during operation. 
Observation is ``a technique that provides a direct way of viewing individuals in their environment performing their jobs or tasks and carrying out processes"~\cite{ISO24765:2017}.
Let us describe a unit test that can be conducted to verify requirement RQ1.5. 
A soft pneumatic gripper with two fingers can be fabricated as proposed in \cite{Partridge2022} using 30\% recycled material. 
One can use black pigments to track the granules and chambers, and the constraining layer's curvature can be tracked using red pigments. 
The fingers can be actuated and a camera can be used to capture their motion. 
The curvature with time can be determined by conducting image processing on the video, and the internal pressure with time can be monitored.

As mentioned previously, the engineering requirements (iii a--d) have been formulated by following the guidance provided in~\cite{NASA2007}. 
On the one hand, most engineering requirements contain numeric tolerances (verifiable). 
For example, RQ1.5 identifies a specific pressure range. 
Similarly, in RQ4.4, we identify a maximum value of 2 N for gripping force when gripping a hard-fragile item. 
On the other hand, there are several high-level requirements which need to be refined into verifiable terms before any verification method can be applied.  
For instance, requirements RQ1.1--1.3 have been formulated at a high-level because the amount of curvature and straightening of a finger can often be dependent on the application. 
In order to be verifiable, we can refine RQ1.1 as: the fingers of the gripper \emph{shall} curve within 2\% of a curve of 10 cm radius when inflated. 
Similarly, we can refine requirement RQ4.2 as: no part of the body of gripping system \emph{shall} have a position equal to the position of the item being grasped. 
This can be verified by monitoring the distance between each of the joints of the gripping system to ensure that there is no collision between the body of the gripping system and the item. 
Meanwhile, with respect to RQ4.1, one does not need to impose an initial velocity on the food item in the test bed, so this requirement can be met by design (i.e. no need to formally verify or monitor it). Thus, the above examples illustrate two versions of the specification for a soft robotic gripper -- one a \emph{verifiable} one from the start (e.g. RQ1.5, RQ4.4); and the other a more \emph{high-level}, unrefined one (e.g. RQ1.1, RQ4.2), for which, through examples, we demonstrate how to make it verifiable. In this manner, this work not only provides a wide-ranging specification for a pick-and-place application, but also provides illustrations of how to formulate verifiable requirements for soft grippers. 

\begin{table}
	\centering
	\caption{\label{Table:Verifiability} Soft-gripper engineering requirements and verification methods.}
	\begin{tabular}{|p{20mm}|p{110mm}|}
		\hline
		\textbf{Requirements} & \textbf{Verification Method} \\ 
		\hline
		RQ1.1,1.2,1.4, 2.1,2.2,2.4 & \emph{Observation} during operation\\
		\hline
		RQ1.3,1.5 & \emph{Practical tests}: unit testing  \\
		\hline
		RQ2.3 & \emph{Practical tests}: edge-case testing \\
		\hline
		RQ2.5,2.6 & \emph{Practical tests}: life-cycle testing \\
		\hline
		RQ3.1,3.2,3.3 & \emph{Practical tests}: repeated testing, and \emph{Observation} \\
		\hline
		RQ4.1 & \emph{Measurement}: vision camera, and \emph{Observation}  \\ 
		\hline
		RQ4.2 & \emph{Measurement}: force torque sensor \\ 
		\hline
		RQ4.3 & \emph{Measurement}: force torque sensor and vision camera  \\ 
		\hline
		RQ4.4 & \emph{Practical tests}: functional test\\ 
		\hline
		RQ4.5 & \emph{Measurement}: displacement sensor   \\ 
		\hline
		RQ4.6 & Both \emph{Practical tests}: functional test, and \emph{Measurement}: displacement sensor \\ 
		\hline
	\end{tabular}
\end{table}

\section{Conclusion} \label{summary-conclusions}
As soft robotics continue to expand into new and diverse industries it is important to consider the design processes used, particularly the development of detailed specifications which can be used to define the requirements for a system and ensure that both functional and non-functional aspects are considered from an early stage of the development process.
Within soft robotics, soft gripping is considered to be one of the most mature areas. 
For wider adoption and acceptability, we must build soft grippers worthy of our trust. 
In this context, this work proposed an extensive \emph{specification} for a soft gripper covering several functional and non-functional properties, including predictability, reliability, adaptability, safety, ethics, and regulations. 
In addition, we promoted the notion of \emph{verifiability} of a soft gripper as a first-class design objective, and provided illustrations of how to formulate verifiable requirements. 
This work was explored using the pick-and-place tasks of grocery items in an automated warehouse. 

We conclude that specifying for trustworthiness in soft robotics as complete systems for real world applications should use a multi-disciplinary approach with inputs from a range of experts including soft roboticists and engineers, as well as experts from the social sciences and humanities.  A multi-disciplinary approach will help ensure that any specification covers both functional and non-functional requirements and will help develop trustworthy soft robots.

\section*{Acknowledgements}
This work has been supported by the UKRI Trustworthy Autonomous Systems Node in Functionality under Grant EP/V026518/1. J.R.\ is supported by EPSRC grants EP/R02961X/1, EP/S026096/1, EP/V062158/1, and EP/T020792/1, and the Royal Academy of Engineering through the Chair in Emerging Technologies scheme, grant CiET17182$\backslash$22. 

\bibliographystyle{unsrt}  
\bibliography{Spec-SoftRobotics-Bib.bib}		
\end{document}